%% file: 0_main.tex
\documentclass{article}

\usepackage{arxiv}

\usepackage[utf8]{inputenc} 
\usepackage[T1]{fontenc}    
\usepackage{hyperref}       
\usepackage{url}            
\usepackage{booktabs}       
\usepackage{amsfonts}       
\usepackage{microtype}      
\usepackage{graphicx}

\usepackage{paralist}
\usepackage{amsfonts}
\usepackage{amsmath}
\usepackage{amssymb}
\usepackage{amsthm}
\usepackage{nameref}
\usepackage{cite}
\usepackage[numbers]{natbib}
\usepackage{algorithm}
\usepackage{tikz}
\usepackage{algpseudocode}
\usepackage{multirow}
\usepackage{newfloat}
\usepackage{listings}
\graphicspath{ {./images/} }
\usepackage{appendix}

\title{Discret2Di - Deep Learning based Discretization for Model-based Diagnosis}

\author{
 Lukas Moddemann, Henrik Sebastian Steude, Alexander Diedrich and Oliver Niggemann \\
 Institute of Automation, Helmut-Schmidt-University, Hamburg, Germany\\
e-mail: surname.name@hsu-hh.de
}

\begin{document}
\maketitle
\begin{abstract}
Consistency-based diagnosis is an established approach to diagnose technical applications, but suffers from significant modeling efforts---especially for dynamic multi-modal time series.
Machine learning seems to be an obvious solution, which becomes less obvious when looking at details: Which notion of consistency can be used? 
If logical calculi are still to be used, how can dynamic time series be transferred into the discrete world? 

This paper presents the methodology Discret2Di for automated learning of logical expressions for consistency-based diagnosis. 
While these logical calculi have advantages by providing a clear notion of consistency, they have the key problem of relying on a discretization of the dynamic system. 
The solution presented combines machine learning from both the time series and the symbolic domain to automate the learning of logical rules for consistency-based diagnosis.
\end{abstract}

\input{1_intro}
\input{2_related_work}
\input{2_1_background}
\input{3_disc2di_approach}
\input{4_evaluation}
\input{5_Discussion}
\input{6_conclusion}

\section*{Acknowledgments}
This research as part of Project (K)ISS is funded by dtec.bw – Digitalization and Technology Research Center of the Bundeswehr. 
dtec.bw is funded by the European Union – NextGenerationEU.
\bibliographystyle{abbrvnat}
\bibliography{0_main}

\appendix
\input{7_appendix}

\end{document}

%% file: 1_intro.tex
\section{Introduction} \label{introduction}
The automated identification of root causes during abnormal system behavior is essential for enhancing operational efficiency, safety protocols, and overall functionality in technical systems.
In the context of evolving environments and technical systems, a prominent challenge arises in the form of continuous system updating and modeling requirements.
Even a minor alteration in the technical system can lead to extensive modeling efforts, consuming valuable time and resources. 
By harnessing the power of Machine Learning (ML), it becomes feasible to efficiently learn new system configurations, enabling a streamlined adaptation process.
However, when addressing Model-Based Diagnosis (MBD), integration of time series data into symbolic logic becomes imperative for learning system models.

Within the realm of MBD, two distinct model types have been specified: Weak-Fault Models (WFM) \citep{de1992characterizing} and Strong-Fault Models (SFM) \citep{struss1989physical}. 
While SFM represent the system in every individual abnormal behavior, WFM characterize the system solely in relation to its non-faulty behavior.
The feasibility of learning a SFM is limited due to the inherent nature of Cyber-Physical Systems (CPS) data, which predominantly consists of observations during normal conditions and lacks comprehensive coverage of all possible faults.
Consequently, the adoption of WFMs for Consistency-Based Diagnosis (CBD) in CPS plays a pivotal role in learning system descriptions.

The challenge lies in obtaining a logical representation of a WFM from time series data and effectively leveraging ML as a potential yet demanding solution for CBD.
To achieve this, three essential steps must be addressed:
\begin{inparaenum}
    \item Discretization of dynamic time series data within normal operating conditions, which encompasses both temporal and numerical dimensions, to extract symbolic state representations.
    \item Obtain a mapping of causal relationships between system components and its symbolic state representations.
    \item Learning of logical rules for the unification of gathered information.
\end{inparaenum}

Symbolic representations are meant to signify a temporal segment of a technical system, comparable to a finite automaton, where a section of a time series represents e.g. the tank filling state within a tank system.
The question arises: how can symbolic representations in case of WFM be derived from time series data, and how can inconsistencies be detected?
\citet{biswas2016approach} presents an approach that centers on anomaly detection utilizing regions of unsupervised learned nominal behavior, which collectively encompass the system's operational behaviors. 
Within our framework, the stringent delineation of operational conditions forces us to refer to these as observational states.
\begin{figure}[ht]
\centering
\includegraphics[width=0.44\textwidth]{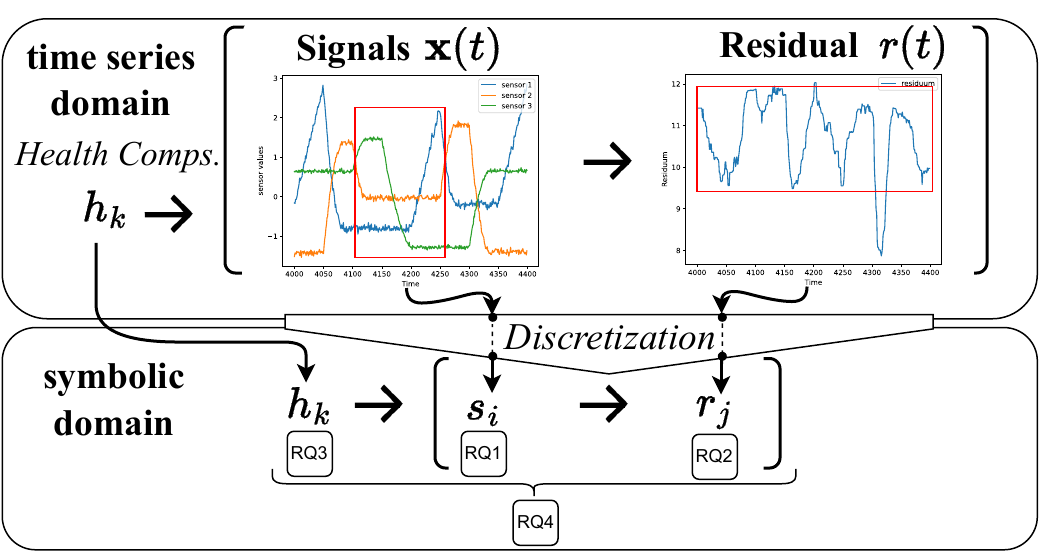}
\caption{
Learning CBD rules as a set of implications to model a WFM along with its Health Components $h_k$ (cf. Def. \ref{def: System Description}). 
$n$ signals from time series data $\mathbf{x}(t) \in \mathbb{R}^n$ are discretized in time and numerical dimensions into observational states $s_i$ (cf. Def. \ref{def: Discretization Observations}) and residual data $r(t) \in \mathbb{R}$ into residual states $r_j$ (cf. Def. \ref{def: Discretization Residuals}), with the red boxes indicating the states.
The research questions (RQs) were identified in the different areas.
}
\label{fig:intro}
\end{figure}

To establish an assertion of a faulty condition within a WFM, an inference from normal system state to a current state must be formulated. 

\textbf{Example:}
Fig. \ref{fig:intro} illustrates an overview of the logical rule framework $h_k \rightarrow (s_i \rightarrow r_j)$.
Here, $h_k$ signifies a conjunction of sensors and actuators that causally influence an observational state $s_i$ (e.g. tank filling).
The observational state's health is assessed by its implication on its discretized residual $r_j$.
In this context, the residual assigns a truth value to the state, considering values near 0 as \textit{True} and \textit{False} otherwise.

Three requirements concerning discretization must be met to effectively employ CBD:
\begin{inparaenum}[(i)]
    \item Incorporate continuous variables, including time, signal amplitudes and its dynamics.
    \item Ensure determinism, guaranteeing a uniquely determined system for any given set of initial conditions of the normal behaving system and identify deviations in non-normal situations.
    \item Identify relevant system states while excluding irrelevant sub-states to avoid unnecessary state generation that do not contribute to the diagnosis.
\end{inparaenum}

The discretization process involved is nontrivial due to the smooth transitions between system states, making it challenging to determine state changes. 
Generative neural networks have demonstrated excellent performance in various unsupervised learning applications \citep{bengio2013representation}.
Therefore, we pose the following research question (RQ) based on the annotations depicted in Fig. \ref{fig:intro}.
\textit{RQ1: Can Generative Neural Networks (GNNs) enable the discretization of multivariate time series data sets into symbolic observation sets for learning CBD rules?}

In CBD, residuals are commonly employed to quantify the deviation between observed and expected behavior, thereby determining the system's consistency.
In order to represent residuals as fault indicators within the framework of symbolic logic, a discretization of residuals becomes essential to achieve a set of implication rules from system states to residual states according to Fig. \ref{fig:intro}.
In the context of RQ1, the following research question arises.
\textit{RQ2: How can residuals from GNNs be obtained and discretized for use in CBD?}

Causal relations cannot be directly derived from data and must be inferred through methods such as randomized control trials that counteract confounding factors \citep{eberhardt2017introduction}.
Conducting such trials can be costly for CPS. 
Hence, the expertise of system experts is crucial in modeling the logical representation of health components (cf. Fig. \ref{fig:intro}) to its observational states, which addresses RQ3: 
\textit{Can we integrate a-priori knowledge about conditioning the correct behavior of health components into the learning algorithm?}

The integration of observational states (RQ1), residual states (RQ2), and a-priori knowledge (RQ3) is imperative to establish one consistent and homogeneous knowledge base, thereby formulating RQ4:
\textit{How can the independently generated information from RQ1 to RQ3 be combined to form a logical rule-base for CBD?}

The contributions of this paper are as follows:
We propose a discretization method for the automated identification of observational states by use of a Categorical Variational Autoencoder (CatVAE), while considering noise and uncertainty.
Secondly, we present an algorithm generating discretized residual states and explore the integration of heuristic knowledge to learn logical rules for CBD.
Overall, we introduce the novel methodology Discret2Di that integrates the learning of states and discretization of residuals for logical rule learning of WFM in CPS. 

%% file: 2_related_work.tex
\section{Related Work} \label{related_work}

The goal of learning a CBD rule set from multivariate data to perform diagnosis touches the research fields of discretization, association learning and CBD.

\textbf{Consistency-based Diagnosis:}
The root cause problem, as defined by  \citet{de1987diagnosing} and \citet{reiter1987theory}, is intended to diagnose multiple symptoms using a WFM.
However, due to the exponential growth of the combinatorial search space, the first proposal by \citet{de1987diagnosing} using the General Diagnostic Engine (GDE) is impractical for huge systems. 
To address this, researchers have investigated the use of Boolean Satisfiability (SAT) solver.
SAT Model-Based Diagnosis algorithms, such as SAFARI \citep{feldman2010solving} and SATbD \citep{metodi2014novel}, transform the diagnosis problem into a Boolean satisfiability problem that can be tackled by SAT solver implementations. 
In contrast, our approach deviates from traditional methods by employing time series data instead of binary circuits.
Though, when it comes to more natural problem descriptions, the Satisfiability Modulo Theory (SMT) solver has been introduced \citep{barrett2018satisfiability}.
\citet{grastien2014diagnosis} presents an SMT-based technique for diagnosing systems of both continuous and discrete data, that breaks down the system into states. 
However, the method utilizes predetermined thresholds for discretization and segments the time dimension based on designated transition times, which  diverges from our unsupervised learning approach.

\textbf{Discretization in Time and Space to model WFM:}
\citet{forbus_qualitative} has published a seminal work describing discretization, utilizing very small time intervals to model continuous behavior in time-discrete systems. 
However, dynamic time series data are complex, necessitating the exploration of non-linear techniques. 
Gaussian Mixture Models (GMMs) \citep{biernacki2000assessing} or k-means clustering \citep{hartigan1979k} are established methods on low dimensional nonlinear data.
Nevertheless, their distance measures are limited to local relations in the data space, and they tend to be ineffective for high-dimensional data \citep{zhang2017deep}.
\citet{hranisavljevic2020discretization} present a method for discretizing multivariate time series into a timed automata representation. 
Their approach involves the utilization of Gaussian-Bernoulli Recurrent Boltzmann Machines to enforce a Bernoulli-distributed data representation.
Moreover, Variational Autoencoder (VAE) \citep{Higgins2017-aw} have become widely popular. 
Nonetheless, creating discrete representations of multivariate data in the latent space is a challenging task for generative networks, owing to the non-differentiability of discrete distributions. 
For this purpose, a continuous reparametrization technique has been introduced based on the  Gumbel-Softmax distribution \citep{jang2016categorical}. 
\citet{fortuin2018som} presented a VAE approach that utilizes a Self Organizing Map as discretization within the latent space. 
In contrast to their methodology, which combines two integrated techniques, our approach demonstrates superiority by minimizing model complexity, utilizing a standard VAE with a modified latent space distribution.

\textbf{Learning of Logical Expressions: }
Detecting co-occurrence patterns among sets of elements is crucial for revealing relationships within a domain.
\citet{agrawal1994fast} proposed the Apriori algorithm for association rule mining.
Still, the approach suffers from the complexity of candidate generation.
To address this, FP-Growth \citep{han2000mining} was introduced, which employs a depth-first search approach and uses the FP-Tree data structure to store frequently occurring information in a compressed form, unlike Apriori's breadth-first search approach.
\citet{kolb2018learning} focuses on learning $\operatorname{SMT}(\mathcal{L} \mathcal{R} \mathcal{A})$ from data, avoiding the reliance on discrete variables by encoding the data into a format amenable to direct solving by SMT solvers.
In contrast, our approach stands out by incorporating pre-discretized data within a dependency relation, effectively converting it into a SAT problem.

To the best of our knowledge, no prior research has introduced a method for learning a CBD rule base for a SAT-solver using a VAE discretization approach that employs a categorical representation.

%% file: 2_1_background.tex
\section{Background} \label{Background}
\newtheorem{definition}{Definition}

CBD faults are located based on the satisfiability of a set of propositional logic expressions. 
In this context, a diagnosis system is a tuple $(SD, COMPS, OBS)$, where $SD$ is the WFM system description in propositional logic, $COMPS$ a set of component symbols, and $OBS$ the observations. 
Consequently, the diagnosis can be defined as follows.

\begin{definition}[Diagnosis]\label{def: diagnosis}
Let $\Delta \subseteq COMPS$.
A diagnosis $D$ for the diagnosis system $(SD, COMPS, OBS)$ is $D(\Delta, COMPS - \Delta)$ and is defined as a conjunction $\left[\bigwedge_{c \in COMPS\setminus \Delta} \neg A B(c) \wedge \bigwedge_{c \in \Delta}  A B(c)\right]$ such that: $SD \cup OBS \cup \{ D(\Delta, COMPS - \Delta )\}$ is satisfiable.
\end{definition} 

To use CBD in conjunction with CPS, the initial step involves discretizing time series signals and residuals into states. 
States serve as symbolic representations with encoded attributes pertaining to the system, with changes occurring over time to capture the inherent dynamics. 

\begin{definition}[Neural Network Discretization of Observational States $S$] \label{def: Discretization Observations}
Let $\mathbf{x}(t)$ be a vector of $n$ sensor values at time $t$. Then  $d_s(\cdot)$ is a discretization function for the extraction of observational states $s_i \in S$, where $S$ is the set of all possible states, such that $d_s(\cdot): \mathbb{R}^n \rightarrow S$, $|S| < \infty\ $.
\end{definition}

Residual values indicate the health of an observational state by comparing sensor values to a model of normal behavior. 
The process of discretizing residuals through the application of thresholds is delineated as follows.

\begin{definition}[Discretization of Residual States $R$]\label{def: Discretization Residuals}
$r(t)$ represents a residual at time $t$, and $d_r(\cdot)$ is a discretization function with parameter $\tau$ for the extraction of residual states $r_j \in R$, where $R$ is the set of all possible residual states, such that $d_r(\cdot): \mathbb{R} \times \mathbf{\tau} \rightarrow R$, $|R| < \infty\ $.
\end{definition}

Subsequently, according to the logical framework illustrated in Fig. \ref{fig:intro}, we investigate how the interconnections between components can be modelled.

\begin{definition}[Health Components $h_k$]\label{def: Health Components}


A health components set \(H = \{h_1, \dots, h_k\}, k \in \mathbb{N}\) consists of components defined as \(h_k = \bigwedge_{c_l \in COMPS_l} \neg AB(c_l)\), where \(c_l \in COMPS_l\) and \(COMPS_l \subset COMPS\). Here, \(COMPS_l\) denotes specific components acting on the system state, and \(\neg AB(c_l)\) denotes the assumption of a healthy \(c_l\).

\end{definition}

However, the description of a system within CBD is essential for automated learning of logical expressions. 
The system description is in this case referred to as a WFM.

\begin{definition}[System Description]\label{def: System Description}
A System Description (SD) of a WFM can be described as a set of implications $h_k \rightarrow \Phi_m$, $k \in \mathbb{N}$, with the logical statement $\Phi_m$, $m \in \mathbb{N}$. 
$\Phi_m$ represents the implication $s_i \rightarrow r_j$, where $s_i \in S$ denotes a observational state of the system in relation to $r_j \in R$, representing the residual state. 
\end{definition} 

%% file: 3_disc2di_approach.tex
\section{Discret2Di Methodology} \label{Methodology}

To automate the extraction of logical expressions from data for use in CBD, a fusion of ML in the time series and the symbolic domain is to be realised.
Existing research has primarily centered on extracting contraints from SMT formulae \citep{kolb2018learning}. 
Nevertheless, ML can offer a potential solution in the application of time series data. 

\subsection{From Discretization to Diagnosis - Methodology}

In the ensuing section, we provide an overview and examine each step of the methodology in context of the RQs.

\begin{figure}[ht]
\centering
\includegraphics[width=0.49\textwidth]{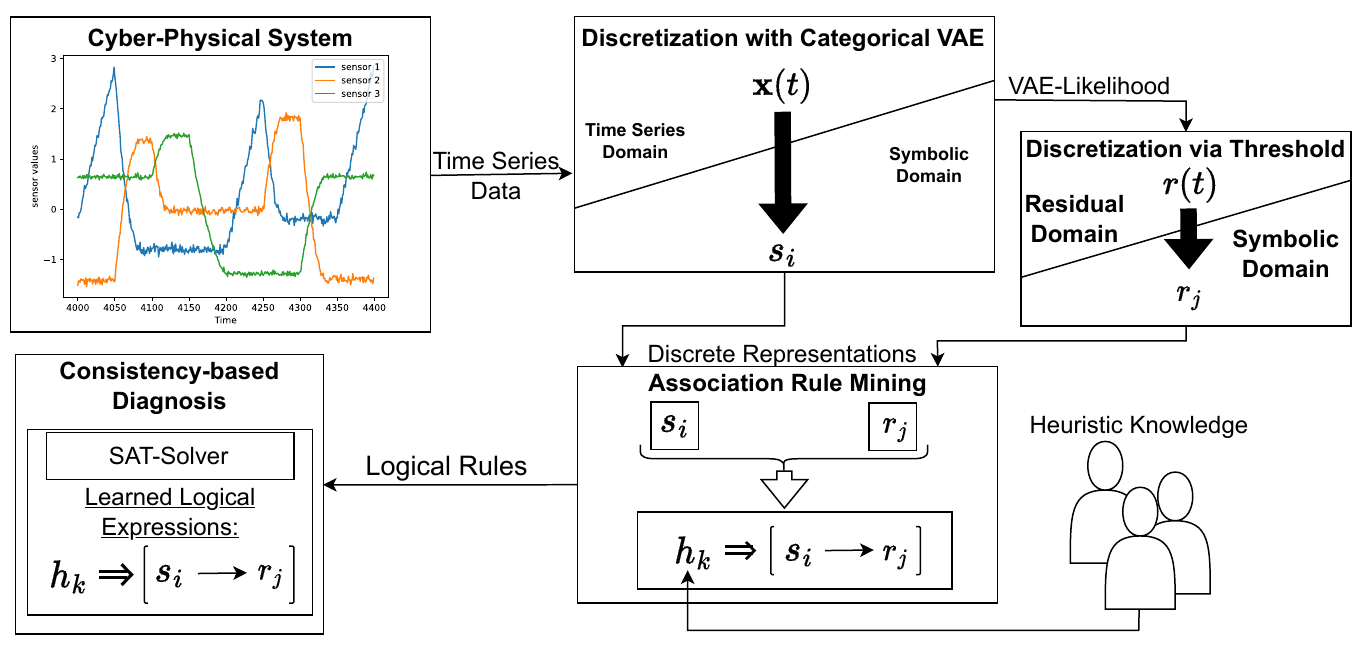}
\caption{Sequence of the methodology Discret2Di. 
By discretizing the CPS data with a CatVAE, which also provides the likelihood as a measure of residual fit, the necessary symbols of $s_i$ and $r_j$ can be obtained. 
By incorporating heuristic knowledge about the Health Components $h_k$, logical expressions can be learned and subsequently solved using a SAT-solver.}
\label{fig:methodology}
\end{figure}

This study proposes the use of a VAE for discretizing multivariate data $\mathbf{x}(t)$ into symbolic observational states $s_i$ (cf. Fig. \ref{fig:methodology}, 'Discretization with Categorical VAE'). 
 Through the integration of a categorical latent space within the VAE framework, an effective categorization of data into discrete states is enabled, thereby facilitating the acquisition of system noise by sampling from the learned probability distribution. 
Nevertheless, caution is required in distinguishing the states precisely, as the use of time windows favours the possibility of an overlap of the states, posing a challenge to their distinct state identification. 
To solve this, the input data for this approach must consist exclusively of measured values at individual points in time.
Residuals $r(t)$ allow the detection of deviations, serving as indicators of anomalies. 
We propose using the CatVAE's likelihood to discretize these residuals into residual states $r_j$ via thresholds (cf. Fig. \ref{fig:methodology}, 'Discretization via Threshold').
Until now, a specific logical relationship between observational states $s_i$ and residual states $r_j$ as outlined in Def. \ref{def: System Description} has not been established.
To address this, we employ an association rule learning approach, generating logical implications (cf. Fig. \ref{fig:methodology}, 'Association Rule Mining'). 
To finalize the formulation of these rules, domain experts with heuristic expertise contribute their insights regarding the health attributes of components.
This approach was  chosen due to the constraint that deriving system knowledge from normal data is not feasible.
The heuristic is solely employed for rule completion, ensuring  a significantly automated rule generation.
Finally, the knowledge learned can be utilized in the form of logical formulas for CBD (cf. Fig. \ref{fig:methodology}, 'Consistency-based Diagnosis'). 

\begin{figure}[ht]
\centering
\includegraphics[width=0.48\textwidth]{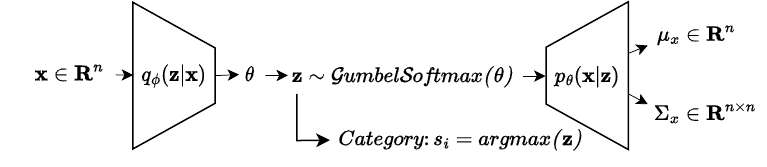}
\caption{CatVAE Architecture.
Input $\mathbf{x}$ is a n-dimensional vector of signals at time $t$. 
The output is a Gaussian probability distribution $p_{\mathbf{\phi}}(\mathbf{x}|z)$, where $z$ is sampled from an approximated categorical distribution obtained by $GumbelSoftmax$. 
During training, we optimize the parameters $\mathbf{\phi}$ and $\mathbf{\theta}$ such that the likelihood of $\mathbf{x}$ under $p_{\mathbf{\theta}}(\mathbf{x}|z)$ is maximized and the KL-Divergence between $q_{\phi}(z|\mathbf{x})$ and the prior $p(z) = \left[\frac{1}{K}, \frac{1}{K}, \ldots,\frac{1}{K}\right]$, over $K \in \mathbb{N}$ categories is minimized. The category $s_i$ is calculated by $argmax$ of $z$.
}
\label{fig:CatVAE}
\end{figure}

\textbf{Discretization of States with Categorical VAE: }
To address RQ1, we revisit the requirements formulated in the \nameref{introduction} for discretizing time series data $\mathbf{x}(t)$ into observational states $s_i$.
Due to the requirement to consider continuous variables such as system dynamics, sensor noise, and the ability to learn meaningful representations from unlabeled data, we employ a probabilistic deep neural network. 
In this approach, we utilize a CatVAE to extract categorical representations $s_i$ from the latent space $\mathbf{z}\in[0, 1]^{K}$, where $\quad \sum_{i=1}^{K} z_i = 1$ and the number of categorical dimensions $K \in \mathbb{N}$ (Fig. \ref{fig:CatVAE}, 'Discretization with Categorical VAE').
Training a CatVAE poses a challenge in terms of discrete sampling and gradient estimation. 
To overcome this, we employ the Gumbel-Softmax relaxation \citep{jang2016categorical}, which provides a continuous relaxation of the discrete distribution, enabling the use of gradient-based optimization algorithms.

The latent space encodes the information in an approximated categorical domain, enforced by the prior $p(\mathbf{z})$ (cf. Eq. \ref{eq: loss function}), whereas the category $s_i$ is obtained by $argmax(\mathbf{z})$.
Since we assume input $\mathbf{x}$ follows a Gaussian distribution, the decoder tries to approximate a probability distribution by determining the mean $\mu_x$ and standard deviations $\Sigma_x$ in each dimension of $\mathbf{x}$ (cf. Fig. \ref{fig:CatVAE}).
To accurately map time series data to latent states, the principles of Kullback-Leibler (KL) divergence and likelihood computation need to be considered in VAE training.
This is balanced using the parameter $\beta$ \citep{higgins2017beta}, which allows fine-tuning between mapping accuracy and latent space structure.
\begin{equation}\label{eq: loss function}
    \mathcal{L}_{VAE}=\mathbb{E}_{q(z|x)}[\text{ln } p(x|z)] - \beta D_{KL}[q(z|x) || p(z)]
\end{equation}
However, the requirement of detecting state transitions accurately is essential and this should be considered when selecting the model's input data. 
To address this, we train the model on time points rather than time sequences to ensure adequate transitions are captured. 

\textbf{Discretization of Residuals via Threshold: }
By leveraging the CatVAE, we can derive the likelihood to assess whether the system behaves normally or not. 
The likelihood serves as residual $r(t)$ of the model to the system input $\mathbf{x}$, providing a probabilistic representation while accounting for the uncertainty associated with the model's estimation.
In our case, the likelihood will be computed using the logarithmic probability of a sample $x$ to the decoder output distribution as represented within the first part of Eq. \ref{eq: loss function}.
To enable statement-making about the possibly non-normal behaving state $s_i$ within real data, the residuals need to be discretized to residual states $r_j$ (cf. Def. \ref{def: Discretization Residuals}) as inconsistency indicator.
Discretization via thresholds is an approach that allows clear decisions. 
Once thresholds are heuristically determined, it becomes straightforward to categorize residuals into discrete states (Def. \ref{def: Discretization Residuals}). 
The process is not restricted to binary categorization but rather allows for the introduction of soft and hard intervention boundaries, leading to the possibility of multiple classifications.
We initially focus on binary discretization, employing a threshold denoted as $\tau$.

\textbf{Association Rule Mining: }
Given the acquired knowledge of the observational states $s_i$ and residual states $r_j$, along with their corresponding timestamps, we are currently positioned to delineate a relationship between these two information resources for the formulation of logical expressions.
The FP-Growth algorithm \citep{han2000mining} is used to create a relation between observational and residual state as required for the WFM system description, where the results will represent $\Phi_m = s_i \rightarrow r_j$ (cf. Def. \ref{def: System Description}).
The algorithm efficiently discovers frequent patterns in the data from a tree structure in a bottom up exploration, allowing for the pruning of rules that fail to meet a specified confidence threshold. 
The main challenge lies in identifying the most valuable rules among the various sources of information. 
Nevertheless, the ability to determine the relevance of a relation offers advantages, as it enables the filtration of potential noise or point anomalies.

\textbf{Heuristic Knowledge Integration: }
To enrich the learned partial rules $[s_i \rightarrow r_j]$ derived during Association Learning (Figure \ref{fig:methodology}, 'Association Learning'), we incorporate heuristic knowledge about Health Components $h_k$. 
The associated observational states are presented to the system expert, who establishes connections between the associated observational state and the causally linked non-observable components of the system according to Def. \ref{def: Health Components}, leading to a WFM of the normal working system (Def. \ref{def: System Description}).
In analogy to a tank example, $h_k$ corresponds to e.g. $ok(pump) \wedge ok(valve)$, where $ok(\cdot)$ can be replaced with $\neg AB(\cdot)$ according to Def. \ref{def: Health Components}, such that a rule with a normal behavior ($r_j = True$) can be formulated as $\neg AB(pump) \wedge \neg AB(valve) \rightarrow (filling_{state} \rightarrow True)$.
By dynamically inferring the identification of observational states $s_i$ and their corresponding residual states $r_j$, we can continuously monitor their health status and indicate inconsistencies.  

\textbf{Consistency-based Diagnosis: }
Once the complete rule base is generated, automated system diagnosis using CBD can be effectively conducted (Def. \ref{def: diagnosis}). 
The learned rules can be treated as a satisfiability problem in the event of a symptom, thus enabling their evaluation through SAT solvers to determine their consistency. 
For this purpose, we employ a SAT solver to compute the unsatisfiable kernel of the inconsistent rule base, encompassing the components that can be regarded as potential root causes.

\subsection{Algorithm to Discret2Di}
In this section, a comprehensive depiction of the algorithmic implementation is provided.
The distinction of the algorithm between learning rules (cf. Alg. \ref{alg:learningRules}) and diagnosing (cf. Alg. \ref{alg:diagnosis}) is based on their respective inputs. 

Initially, Algorithm \ref{alg:learningRules} is executed to learn rules of the normal system behavior. 
This procedure entails generating information concerning observable states and their corresponding likelihoods using a CatVAE, trained on each individual timestamp $\mathbf{x}_t$ within the entire dataset $\mathbf{x}_{norm}$.
The CatVAE is designed specifically for categorizing data in the latent space, where it has a categorical distribution as a prior. 
By using a temperature parameter, the categorical distribution can be approximated and gradients can be computed during training using the reparameterization trick. 
In this study, the temperature parameter is kept low at $0.5$, as the expected value of the Gumbel-Softmax random variable approaches the expected categorical value with the same logits \citep{jang2016categorical}. 
Elevating the value would result in a uniform distribution over the prior.
\algnewcommand\algorithmicforeach{\textbf{for each}}
\algdef{S}[FOR]{ForEach}[1]{\algorithmicforeach\ #1\ \algorithmicdo}

\begin{algorithm}[ht]
\caption{Pre2Di}\label{alg:learningRules}
\begin{algorithmic}[1]
\renewcommand{\algorithmicrequire}{\textbf{Input:}}
\renewcommand{\algorithmicensure}{\textbf{Output:}}
\Require $CatVAE_{model}$, $\mathbf{x}_{norm}$, $\tau_{a}$, $h_k$
\Ensure $\Phi$

\ForEach{$\text{row} \; \mathbf{x_t} \; \text{of} \; \mathbf{x}_{norm}$}

\State $likelihood, z_{logits} \gets CatVAE_{model}(\mathbf{x}_t)$
\State $S \gets S + argmax(z_{logits})$ \Comment{Def. \ref{def: Discretization Observations}}
\State $R \gets R + d_r(likelihood)$ \Comment{Def. \ref{def: Discretization Residuals}}
\EndFor

\State $\Phi_{partial} \gets AssociationRules(S, R, \tau_a)$ 
\ForEach {$\varphi_{partial} \in \Phi_{partial}$}
\State $\varphi_i \gets (h_k \rightarrow \varphi_{partial})$ \Comment{Def. \ref{def: System Description}}
\EndFor
\State $ \Phi \gets \bigcup_{i=1}^n \varphi_i$
\end{algorithmic}
\end{algorithm} 

Moreover, a parameter $\beta$ (cf. Eq. \ref{eq: loss function}) is introduced to scale the KL-divergence during training of the CatVAE. This enables the training loss to be adjusted with regard to the reconstruction of input data (cf. Fig. \ref{fig:CatVAE}).
Upon acquiring the CatVAE ($CatVAE_{model}$), the algorithm computes the latent approximated categorical logits and likelihood (cf. Eq. \ref{eq: loss function}) for each timestamp $\mathbf{x}_t$ within the input data $\mathbf{x}_{norm}$ (cf. Alg. \ref{alg:learningRules}, line 1-2).
To meet the requirement of deterministic generation of observational system states, sampling within the latent space is deactivated after the model training.
By determining the $argmax()$ of the categorical logits $z_{logits}$, the observational state $s_i$ can be derived and  added to a persistent list $S$ (cf. Alg. \ref{alg:learningRules}, line 3). 
Following this, the residuals are subjected to binary discretization function $d_r$ by employing a heuristically defined threshold based on the likelihood (cf. Alg. \ref{alg:learningRules}, line 4). 
The function $AssociationRules(\cdot)$ requires inputs such as the observational states set $S$, the residual states set $R$, and a threshold $\tau_{a}$, which serves as a metric to determine the relevance of a candidate rule represented as $s_i \rightarrow r_j$. 
Once obtained the candidate rules $\Phi_{partial}$ (cf. Alg. \ref{alg:learningRules}, line 6), our approach encompasses the integration of health components $h_k$ from the observable state (cf. Alg. \ref{alg:learningRules}, line 7-9). 
This involves adding the respective components for each rule with the help of system experts.
The resulting output $\Phi$ represents the system rule base in form of a WFM (cf. Alg. \ref{alg:learningRules}, line 10).

\begin{algorithm}[ht]
\caption{Discret2Di}\label{alg:diagnosis}
\begin{algorithmic}[1]
\renewcommand{\algorithmicrequire}{\textbf{Input:}}
\renewcommand{\algorithmicensure}{\textbf{Output:}}
\Require $CatVAE_{model}$, $\mathbf{x}_{anom}$, $\mathbf{x}_{norm}$, $\tau_{a}$,  $h_k$
\Ensure $\delta$

\State $\Phi \gets Pre2Di(CatVAE_{model}, \mathbf{x}_{norm}, \tau_{a}, h_k)$  \Comment{Alg. \ref{alg:learningRules}}
\ForEach{$\text{row} \; \mathbf{x_t} \; \text{of} \; \mathbf{x}_{anom}$}

\State $likelihood, z_{logits} \gets CatVAE_{model}(\mathbf{x}_{t})$ 
\State $S \gets argmax(z_{logits})$ \Comment{Def. \ref{def: Discretization Observations}}
\State $R \gets d_r(likelihood)$ \Comment{Def. \ref{def: Discretization Residuals}}
\State $\delta \gets diagnose(\Phi, S, R)$  \Comment{Def. \ref{def: diagnosis}, Def. \ref{def: System Description}}
\EndFor
\end{algorithmic}
\end{algorithm}

The trained $CatVAE_{model}$, along with anomalous data $\mathbf{x}_{anom}$, serve as input to Algorithm \ref{alg:diagnosis} to perform CBD. 
The categorical logits $z_{logits}$ and their respective likelihoods are generated based on the individual timestamps $\mathbf{x}_t$ of $\mathbf{x}_{anom}$ (cf. Alg. \ref{alg:diagnosis}, line 2-3). 
Consequently, the resulting outputs are the discretized representations of $z_{logits}$ and likelihoods.
The diagnosis function $diagnose(\cdot)$ receives inputs such as the rule base $\Phi$ (cf. Alg. \ref{alg:diagnosis}, line 1), discretized observational states set $S$, and residual states set $R$, wherein $\Phi$ can be equated with the system description (SD), and $S$ and $R$ with the observations (OBS).
In case of an inconsistency in the logical rule base, a diagnosis $\delta$ with its root-cause will be computed (cf. Alg. \ref{alg:diagnosis}, line 6). 

%% file: 4_evaluation.tex
\section{Empirical Evaluation} \label{Evaluation}
\newtheorem{property}{Property}

We evaluated Discret2Di on various data sets.
Addressing RQ1, we describe two artificial and two real data sets to demonstrate the efficacy of the CatVAE discretization and residual generation compared to Gaussian Mixture Models (GMMs).
In order to investigate research questions RQ2 and RQ3, we employed a simulated data set representing a three-tank system with different fault scenarios. 

\subsection{Evaluation of CatVAE Discretization}

In this section, we assume that CPS data in a static sense follow a specific probability distribution. 
Because of the variety of CPS, we theoretically discuss the power of the CatVAE with respect to two different system properties. 

\begin{property}[Multiple Modes]
Often, automation data will show multi modal behavior due to unobserved discrete variables. 
We investigate a simple mixture of three Gaussian components where each component is $\mathbf{x} \sim \mathcal{N}(\mu, \sigma^2)$ and $\sigma^2$ represents the diagonal covariance matrix.
\label{propierty:multiplemode}
\end{property}

\begin{property}[Multiple Modes and Redundant Variables]
Another characteristic of CPS are redundant variables, where there is a lower-dimensional latent representation to the $N_x$-dimensional data.  
We investigate data that is distributed on an elliptic curve with a small Gaussian noise component in combination with multiple modes.
\label{propierty:combined}
\end{property}

\begin{figure}[ht]
\centering
\begin{tikzpicture}
    \draw (0, 9) node[inner sep=0] {\includegraphics[width=0.19\textwidth]{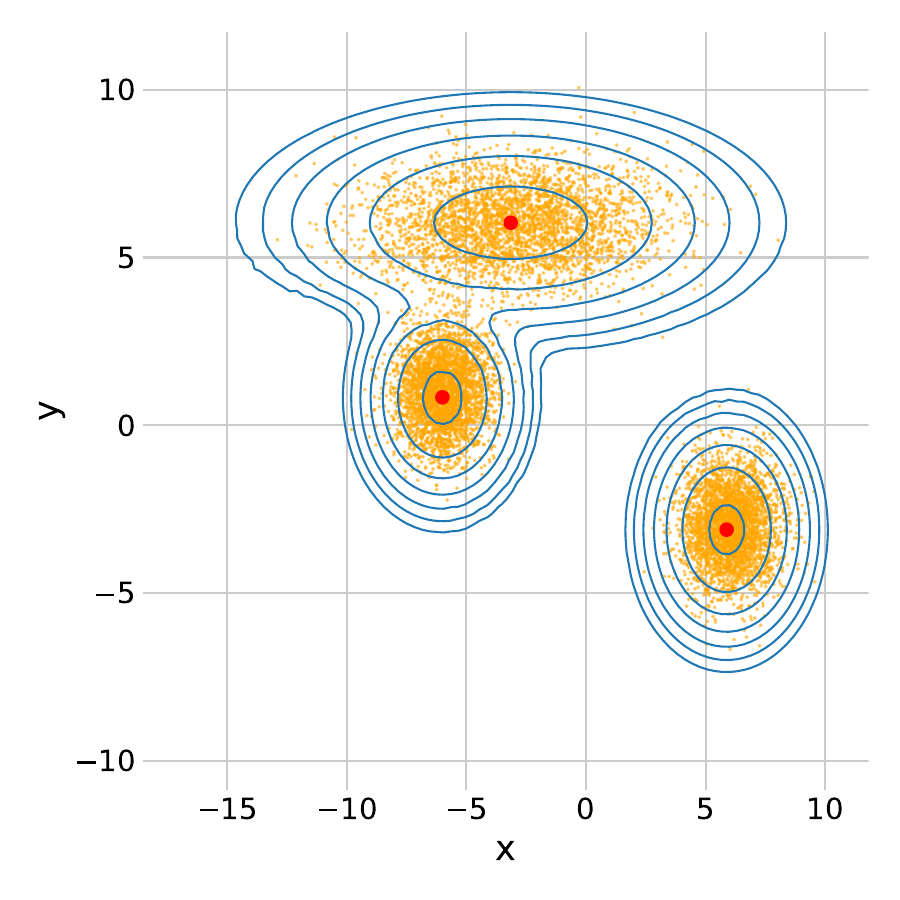}};
    \draw (4.5, 9) node[inner sep=0] {\includegraphics[width=0.19\textwidth]{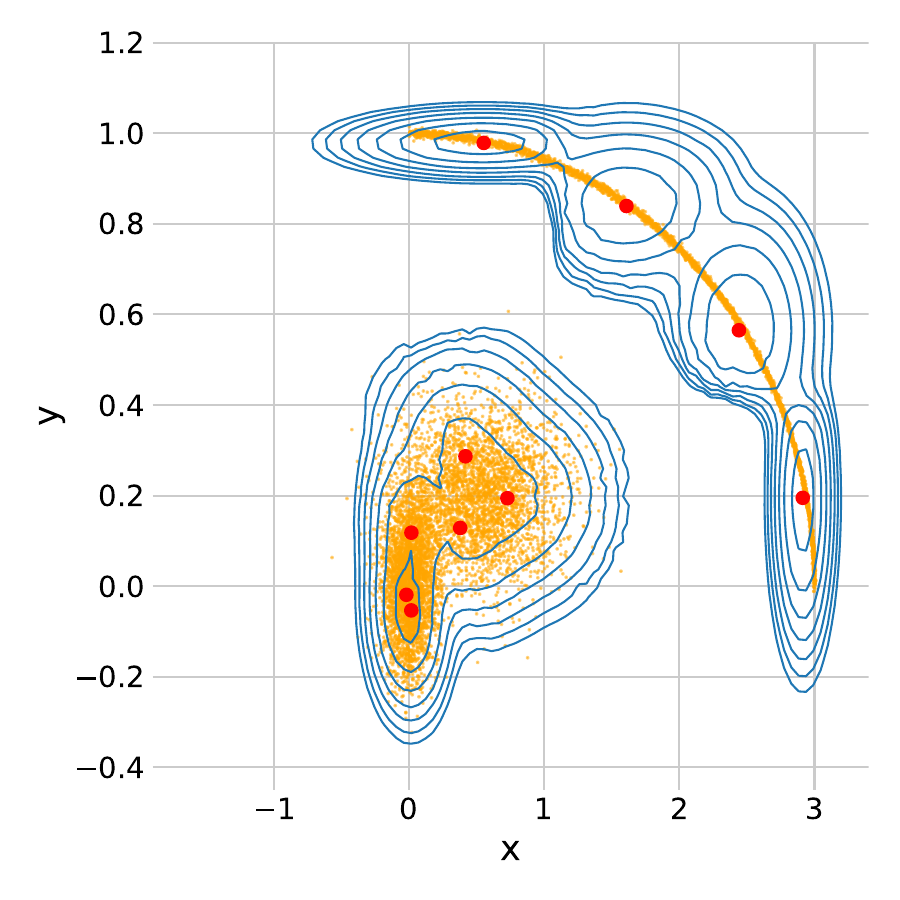}};
\end{tikzpicture}
\caption{Datasets for multiple modes and redundant variables (orange points). Presentation of discretized states (red points) and their according likelihood levels (blue lines)}\label{fig:sysa123}
\end{figure}

To validate the algorithm's ability to handle Property \ref{propierty:multiplemode}, we generated a dataset with three modes using different distributions. 
The second dataset consists of two modes that are closer to each other in the 2D data space, with a redundant variable represented by an ellipse (cf. Fig. \ref{fig:sysa123}).
To assess the CatVAE's efficacy in discretizing time series data, we visualize the learned categories as dots, along with their associated likelihoods.
The findings pertaining to Property 1 demonstrate the effectiveness of data discretization through the utilization of three distinct categories.
In the context of Property 2, we granted the CatVAE a degree of freedom of 10 categories in the latent space. The outcomes (cf. Fig. \ref{fig:sysa123}) revealed that, when dealing with multiple modes, the categories tend to accumulate within the respective modes.
This observation suggests that the system states can be approximated, leading to the discovery of an underlying structure, where certain categories exhibit similar characteristics.

However, during rule learning, no rule emerges that cannot be represented by plausible system descriptions with corresponding system states. 
Moreover, based on Property \ref{propierty:combined}, the CatVAE can approximate the shape of the system and learn the underlying system dependencies.

To account for the extraction of categories and its residuals, we employ a comparison of the CatVAE with a GMM as baseline using the BeRfiPl simulated modular process engineering dataset \citep{ehrhardt2022ai} and two real datasets from process engineering. 
One real dataset is the SWaT dataset \citep{mathur2016swat}, where we will focus on process P1. 
The other real dataset relies on the Smart Automation (SmA) research plant by Siemens, representing a four tank batch process plant.\footnote{Siemens Dataset of a Four-Tank Batch Process: https://github.com/thomasbierweiler/FaultsOf4-TankBatchProcess}
Below, we have conducted experiments to demonstrate the strengths and weaknesses of the algorithms to discretization and the usefullness of the residuals for anomaly detection.

\begin{table}[ht]
  \caption{Comparing CatVAE and GMM for discretization and residual generation.}
    \centering
\scalebox{0.75}{
\begin{tabular}{c|c|c c c}
    \centering
\multirow{2}{*}{\textbf{Algorithm}} & \multirow{2}{*}{\textbf{Discipline}} & 
\multicolumn{3}{c}{\textbf{Datasets}}\\
& & BeRfiPl & SWaT & SmA\\
\hline \hline
CatVAE  & Discretization & \checkmark & \checkmark & \checkmark \\
 & Residual Generation & x & \checkmark & x\\
\hline
GMM  & Discretization & \checkmark & x & x \\
 & Residual Generation & \checkmark & \checkmark & \checkmark\\
\end{tabular}
}
\end{table}

Although CatVAE's performance in residual extraction may currently fall short, its advantage lies in its independence of a priori knowledge regarding the system's state quantity—a requisite of GMM. 
The discrepancy in performance can be attributed to a pivotal contrast in covariance matrix treatment: CatVAE adopts a diagonal covariance matrix, whereas GMM embraces a full covariance matrix strategy.
Conversely, the GMM determines the quantity of categories, yet it does not consistently find the most significant categories, as exemplified by CatVAE.
\subsection{Evaluation of Discret2Di}
To evaluate our methodology Discret2Di comprehensively, we utilized a simulation of a three-tank model \citep{steude2022learning}, where the heuristic knowledge is a priori known.
The simulation is based on an inflow, observable valves and fill levels. 
To evaluate our method, we varied the signal-to-noise ratio and the length of the introduced anomalies to determine at which point an anomaly becomes significant and whether the diagnosis remains consistent with the true root cause.

In Table \ref{tab:three tank dataset}, the parameters according to the simulation are specified. 

\begin{table}[ht]
    \caption{Simulation setup for three-tank dataset. Each parameter describes the mode of operation, where 1 in the case of $q_i$ means full pumping power. In the case of valves $kv_i$, 1 corresponds to fully open and 0 to the closed valve.}
    \centering
\scalebox{0.8}{
    \begin{tabular}{c | c c c c c} 
     state & $q1$ & $q3$ & $kv1$ & $kv2$ & $kv3$\\ [0.1ex] 
     \hline\hline
     $Q1$ & 0.1 & 0 & 0 & 0 & 0 \\ 
     \hline
     $V12$ & 0 & 0 & 0.1 & 0 & 0 \\
     \hline
     $V23$ & 0 & 0 & 0 & 0.1 & 0 \\
     \hline
     $V3$ & 0 & 0 & 0 & 0 & 0.1 \\
     \hline
     $V12_{faulty}$ & 0 & 0 & 0.01 & 0 & 0 \\
     \hline
     $V23_{faulty}$ & 0 & 0 & 0 & 0 & 0 \\
     \hline
     $V3_{faulty}$ & 0 & 0 & 0 & 0 & 0.05 \\
     \hline
     $Q1_{faulty}$ & 1 & 0 & 0 & 0 & 0 \\
    \end{tabular}}
    \label{tab:three tank dataset}
\end{table}

The simulations were conducted by simulating the system state for a duration of 50 samples per state, and generating the following sequences with a total of 15 cycles of simulation for the normal state $S_{normal}$. 
We simulated one cycle and three cycles of the anomaly states $S_{anom_i}$, respectively, while maintaining four normal cycles $S_{normal}$ both before and after the anomaly simulation.

\begin{equation}\label{equation: normal simulation}
\small{S_{normal} = [Q1, V12, V23, Q1, V12, V23, V3]}
\end{equation}

Using the normal scenario, we initially trained the proposed methodology in order to extract meaningful observational states and create a logical rule base.
Heuristic knowledge regarding non-observable components, such as $(q1 \wedge v12)$ for the $filling$ state of tank 1, was incorporated into the partially learned rule base as $q1 \wedge v12 \rightarrow (filling \rightarrow (|res|< \tau_{res}))$, where $res$ represents the actual residual and $\tau_{res}$ some heuristic defined threshold to indicate anomalies for the observational state $filling$.
Upon the completion of the rules, the anomaly scenarios were introduced into the method to generate diagnoses (cf. Table \ref{tab:diagnosis}).
Four anomaly scenarios are considered, each involving the replacement of a single faulty state specified in Table \ref{tab:three tank dataset} within the normal sequence behavior from Eq. \ref{equation: normal simulation}.
Analyzing inference data only after the completion of the third cycle is crucial to consider the settling state of the simulation system. 
Moreover, the CatVAE's ability to learn accurate observational states diminishes when exposed to high levels of white noise ($> 0.3 \sigma$). 
This result is expected as higher variances in the learned categories may disregard intermediate observational states, such as the $filling$ state between \textit{tankfull} and \textit{tankempty} states. 
Our evaluation was conducted using a white noise level of $0.1 \sigma$.

\begin{table}[ht]
  \caption{The results for Discret2Di with one and three cycles of anomaly state simulation.}
    \centering
\scalebox{0.75}{
    \begin{tabular}{c | c c c c} 
     $\#$ Anomalous Cycles & $S_{q1anom}$ & $S_{v12anom}$ & $S_{v23anom}$ & $S_{v3anom}$ \\ [0.1ex] 
     \hline
     \hline
     $1$ & \checkmark & \checkmark & x & x \\
     \hline
     $3$ & \checkmark & \checkmark & \checkmark & \checkmark \\
    \end{tabular}}
    \label{tab:diagnosis}
\end{table}

Based on the simulations, it is evident that the detection of anomalies depends on certain time threshold and the severity of the anomaly, as it is the case in the detection and diagnosis of $S_{v23anom}$ and $S_{v3anom}$ for only one anomaly state cycle simulation (cf. Table \ref{tab:diagnosis}). 
However, this also corresponds to the expectations, since the discretization of the health state is created via the likelihood of the respective categories according to the learned $\mu$ and $\sigma$. 
Therefore, if an anomaly is introduced into the system but the effect is too small to exceed the corresponding uncertainty range for the associated category, the anomaly will not be detected and diagnosed.

%% file: 5_Discussion.tex
\section{Discussion} \label{discussion}
Our methodology was initially tested on the primary bottleneck, the CatVAE, to evaluate its effectiveness in discretization.
The results demonstrated that the methodology effectively meets the requirements defined in the \nameref{introduction}.
Through the utilization of real data and generated data from Property 1, we successfully demonstrated the precise categorization using the CatVAE, even in cases involving redundant and correlated system variables. 
Additionally, the application of Discret2DI to a simulated tank model has demonstrated that the methodology can be employed for the automated generation of diagnoses using time-series data.

However, certain gaps were observed in the detection of health status. 
Regarding the discretization of observational states, our findings indicated that, by considering appropriate parameters, the generated symbolized representations offer satisfactory depictions of the system. 
It is important to note that a certain tolerance for learned values must be considered when incorporating categories into a rule for clarity.


%% file: 6_conclusion.tex
\section{Conclusion and Future Work} \label{conclusion}
This paper introduces a novel methodology Discret2Di that integrates time series and symbolic ML techniques for learning logical expressions for the use in CBD. 
We identified four different RQs and examined them for integration within the methodology.
This study effectively demonstrates the efficacy of CatVAEs, upon which our discretization approach is founded. 
Moreover, we provide evidence that the algorithm takes into account the requirements outlined in the \nameref{introduction}, as indicated by the results obtained in the evaluation. 
By incorporating heuristic knowledge, we establish that logical expressions in accordance with the system description within CBD can be derived. 
This approach reduces the reliance on manual modeling techniques, making the overall process more efficient.
A key objective of this approach is to minimize reliance on external information, including system experts, to the greatest extent possible.

To achieve this goal, further investigation into anomaly detection as well as into the causalities among system components, such as the data-driven identification of causalities, is required to enhance the methodology's capabilities.

%% file: 7_appendix.tex
\clearpage
\appendixpage
All experiments were conducted using Python 3.9 on a 64-bit Linux VM equipped with an 8GB Nvidia V100 GPU and a CPU featuring 10 cores. The necessary software package prerequisites are detailed within the repository of the provided code.

In order to establish the replicability of our assessment concerning Discret2Di, the subsequent hyperparameters were employed.

\begin{table}[ht]
\centering
    \caption{CatVAE hyperparameters used for Siemens SmA discretization. The model is trained for 400 epochs. $beta$ and temperature $\tau$ were changed from 0.1 to 1 in 0.1 steps.}
\begin{tabular}{l|l}
Parameter & Values \\
\hline \hline
kernel size & 12 \\ \hline
batch size & 128 \\  \hline
temperature $\tau$ & 0.5 \\ \hline
$\beta$ & 0.1 \\ \hline
early stopping patience & 40 \\  \hline
training / validation samples & 29,740 / 7,436\\ \hline
encoder block (our model)   & (39, 512, 512, 4) \\ \hline
linear layer (encoder - latent) & (4, 20) \\ \hline
decoder block 2 (our model) & (20, 512, 512, 8) \\ \hline
linear layer (decoder - $\mu$) & (8, 39) \\ \hline
linear layer (decoder - $\sigma$) & (8, 39)
\end{tabular}
\label{tab:hyperparameters_SmA}
\end{table}
\begin{table}[ht]
\centering
    \caption{CatVAE hyperparameters used for BeRfiPl dataset \cite{ehrhardt2022ai} discretization. The model is trained for 400 epochs. $beta$ and temperature $\tau$ were changed from 0.1 to 1 in 0.1 steps.}
\begin{tabular}{l|l}
Parameter & Values \\
\hline \hline
kernel size & 12 \\ \hline
batch size & 32 \\  \hline
temperature $\tau$ & 0.5 \\ \hline
$\beta$ & 0.1 \\ \hline
early stopping patience & 40 \\  \hline
training / validation samples & 3,385 / 847 \\ \hline
encoder block (our model)   & (154, 512, 512, 4) \\ \hline
linear layer (encoder - latent) & (4, 20) \\ \hline
decoder block 2 (our model) & (20, 512, 512, 8) \\ \hline
linear layer (decoder - $\mu$) & (8, 154) \\ \hline
linear layer (decoder - $\sigma$) & (8, 154)
\end{tabular}
\label{tab:hyperparameters_BeRfiPl}
\end{table}

\begin{table}[ht]
\centering
    \caption{CatVAE hyperparameters used for SWaT dataset discretization. The model is trained for 400 epochs. $beta$ and temperature $\tau$ were changed from 0.1 to 1 in 0.1 steps.}
\begin{tabular}{l|l}
Parameter & Values \\
\hline \hline
kernel size & 12 \\ \hline
batch size & 128 \\  \hline
temperature $\tau$ & 0.5 \\ \hline
$\beta$ & 0.1 \\ \hline
early stopping patience & 40 \\  \hline
training / validation samples & 48,000 / 12,000 \\ \hline
encoder block (our model)   & (4, 512, 512, 8) \\ \hline
linear layer (encoder - latent) & (8, 20) \\ \hline
decoder block 2 (our model) & (20, 512, 512, 16) \\ \hline
linear layer (decoder - $\mu$) & (16, 4) \\ \hline
linear layer (decoder - $\sigma$) & (16, 4) 
\end{tabular}
\label{tab:hyperparameters_SWaT}
\end{table}

\begin{table}[ht]
\centering
    \caption{CatVAE hyperparameters used for Three-Tank Dataset discretization.  The model is trained for 400 epochs. $beta$ and temperature $\tau$ were changed from 0.1 to 1 in 0.1 steps.}
\begin{tabular}{l|l}
Parameter & Values \\
\hline \hline
kernel size & 12 \\ \hline
batch size & 64 \\  \hline
temperature $\tau$ & 0.3 \\ \hline
$\beta$ & 0.2 \\ \hline
early stopping patience & 40 \\  \hline
training / validation samples & 3,500 / 1500\\ \hline
encoder block (our model)   & (3, 256, 256, 8) \\ \hline
linear layer (encoder - latent) & (8, 10) \\ \hline
decoder block 2 (our model) & (10, 256, 256, 16) \\ \hline
linear layer (decoder - $\mu$) & (16, 3) \\ \hline
linear layer (decoder - $\sigma$) & (16, 3) 
\end{tabular}
\label{tab:hyperparameters_tank}
\end{table}

\begin{table}[ht]
\centering
    \caption{GMM hyperparameters employed for the process of discretization delineated across the used datasets (library: sklearn).}
\begin{tabular}{l|l}
Parameter & Values \\
\hline \hline
random state & 0 \\ \hline
covariance type & "full" \\ \hline
SWaT dataset: n components & 5 \\ \hline
SmA dataset: n components & 4 \\ \hline
BeRfiPl dataset: n components & 7 
\end{tabular}
\label{tab:hyperparameters_gmm}
\end{table}

\begin{table}[ht]
\centering
    \caption{Hyperparameters used for association rule learning by FP-Growth (library: mlxtend).}
\vspace{2mm}
\begin{tabular}{l|l}
Parameter & Values \\
\hline \hline
threshold metric (association learning) & "confidence" \\ \hline
min threshold (association learning) & 0.005 \\ \hline
threshold likelihood & -50 \\ \hline
\end{tabular}
\label{tab:hyperparameters_associationlearning}
\end{table}